\documentclass[conference]{IEEEtran}
\IEEEoverridecommandlockouts

\usepackage{amssymb,amsfonts}
\usepackage{algorithmic}
\usepackage{comment}
\usepackage{textcomp}
\def\BibTeX{{\rm B\kern-.05em{\sc i\kern-.025em b}\kern-.08em
    T\kern-.1667em\lower.7ex\hbox{E}\kern-.125emX}}

\usepackage{cite}
\usepackage{graphicx}
\usepackage[numbers,sort&compress]{natbib}
\usepackage[cmex10]{amsmath}
\usepackage{array}
\usepackage{color}
\usepackage{xcolor}
\usepackage{dirtytalk}
\usepackage{amssymb}
\usepackage{balance}
\usepackage{booktabs}
\usepackage[dvipsnames]{xcolor}
\usepackage{physics}
\usepackage{float}
\usepackage{mathrsfs}
\usepackage{pgfplots}
\usepackage{subcaption}
\usepackage{svg}
\usepackage{circuitikz}
\usepackage{mathtools}
\usepackage{siunitx}
\usepackage{flushend}
\definecolor{mintbg}{rgb}{.63,.79,.95}
\colorlet{lightmintbg}{mintbg!40!white}
\usetikzlibrary{arrows.meta, positioning}

\DeclareMathAlphabet{\mathpzc}{OT1}{pzc}{m}{it}

\usepackage{tikz}
    \tikzset{
    bar/.pic={
        \fill (-.1,0) rectangle (.1,#1) (0,#1) node[above]{\bfseries $#1$};
    }
}
\begin{document}
\newcommand*{\Scale}[2][4]{\scalebox{#1}{$#2$}}%
\newcommand\scalemath[2]{\scalebox{#1}{\mbox{\ensuremath{\displaystyle #2}}}}
\title{Graph Coloring Approach to Solving Sudoku\\ with Oscillatory Neural Networks
\thanks{This work has received funding from the European Union's Horizon Europe research and innovation programme, PHASTRAC project under grant agreement No 101092096 as well as European Research Council ERC THERMODON project under grant agreement No. 101125031. }
}

\author{\IEEEauthorblockN{Filip Sabo and Aida Todri-Sanial}
\IEEEauthorblockA{\textit{NanoComputing Research Lab, Electrical Engineering Department} \\
\textit{Eindhoven University of Technology}\\
Eindhoven, the Netherlands \\
f.sabo@tue.nl, a.todri.sanial@tue.nl}
}

\maketitle

\begin{abstract}
Oscillatory Neural Networks (ONNs) present an attractive physics-based computing paradigm rooted in the dynamics of a network of typically fully coupled oscillators aiming to minimize an underlying energy function. In this paper, we propose an ONN-based solver for one well-known constrained combinatorial optimization problem, namely a Sudoku, by formulating the problem as a Graph Coloring problem. By modifying the already existing Graph Coloring solver to a computationally cheaper version and introducing an additional term ensuring the fulfillment of the Sudoku constraints, our solver is shown to significantly outperform the existing HNN- and ONN solvers in terms of accuracy. In particular, we are able to achieve nearly flawless accuracies on $4 \times 4$ as well as rather high accuracies on $9 \times 9$ Sudoku puzzles for different numbers of unknown digits.
\end{abstract}


\section{Introduction}

Constrained combinatorial optimization problems (COPs) are ubiquitous in the industry, ranging from scheduling through chip design to the finance sector \cite{yang2011computational, yildiz2009effective, fathi2019optimization, bangert2012optimization}. In particular, NP-hard problems present a special subclass of COPs where the time to find the solution becomes exponential as the problem size increases \cite{karp2009reducibility}. Additionally, the current computing paradigm is not suitable for dealing with such problems, as it suffers from the memory–compute bottleneck resulting in high power consumption \cite{von1993first, efnusheva2017survey, Toews.2020}. In this regard, there has recently been a push for alternative computing paradigms. In this paper, we investigate the performance of a physical computing paradigm based on a network of coupled oscillators, called Oscillatory Neural Networks (ONNs) \cite{todri2024computing}.

In contrast to conventional Artificial Neural Networks \cite{rosenblatt1958perceptron, yegnanarayana2009artificial, zou2009overview}, Oscillatory Neural Networks present an emerging physics-based computing paradigm rooted in the dynamics of a network of typically fully coupled oscillators aiming to minimize an underlying energy function \cite{izhikevich2003simple, wang2019oim, todri2024computing}. In such networks, the information is encoded in the phases of the oscillators. Thanks to their close ties to Hopfield Neural Networks \cite{Hopfield1982, ramsauer2021hopfield} and the Ising model \cite{Ising_1925, lucas2014ising}, ONNs are well suited for auto-associative memory tasks \cite{Hoppensteadt_2000, 2026_Haverkort_overcoming_quadratic_digital_onn, abernot2021digital, maher2024implementation, gower2025learningspeedphysicsequilibrium, sabo2024classonn, abernot2023training, Rohan2025Deep, miyato2025artificialkuramotooscillatoryneurons} and combinatorial optimization problems \cite{wang2019oim, haverkort2025solving, sun_general_2025,gonul2025gpuacceleratedsimulatedoscillatorisingpotts, vadlamani_combinatorial_2024, bashar2024fpga,maher_cmos_compatible_2024, kanao_simulated_2022, mallick2022computational}. In this paper, we propose an ONN-based solver for one well-known constrained combinatorial optimization problem, namely a Sudoku.

Sudokus embody the renowned digit placement puzzle. In particular, given a grid of size $N\times N$ and some known digits, the aim lies in filling out the remaining empty cells of the grid with integers from the range $[1,N]$ such that no repetition of digits occurs in any row, column, or box of the grid. Thus, Sudokus can be viewed as a constrained combinatorial optimization problem. In particular, Graph Coloring, the problem of assigning each node in a graph a \say{color} such that a minimal number of colors is utilized and no two adjacent nodes share the same color, offers a well-suited fit for Sudokus. In this paper, we build an ONN-solver tailored to Sudoku's constraints based on the Graph Coloring problem. 

Recently, the authors in \cite{haverkort2025solving} have developed an ONN-based Sudoku solver by embedding the given Sudoku constraints into the weight matrix. Their method has been shown to improve upon the existing HNN-based Sudoku solver for $4\times 4$, $9\times 9$ as well as $16\times 16$ puzzles in terms of accuracy \cite{mladenov2011solving}. In contrast, in this paper, we present an alternative ONN-Sudoku solver based on Graph Coloring with an additional term ensuring the fulfillment of the constraints. Our model is found to significantly outperform the existing HNN- and ONN-solver in terms of accuracy on $4\times 4$ and $9\times 9$ Sudoku puzzles by reaching almost perfect accuracies on $4\times 4$ and predominantly high accuracies on $9\times 9$ Sudokus for various numbers of unknown digits.

This paper is structured as follows. In Section \ref{sec: background}, we introduce the key concepts behind physics-based computing with Oscillatory Neural Networks (ONNs). Furthermore, Section \ref{sec: methods} outlines our proposed ONN-Sudoku solver based on Graph Coloring together with the benchmarking set-up. Most importantly, our benchmarks on $4\times 4$ and $9\times 9$ Sudokus in terms of accuracy against the established HNN- \cite{mladenov2011solving} and ONN-Sudoku solvers \cite{haverkort2025solving} are presented in Section \ref{sec: results}. Last but certainly not least, the results are discussed in Section \ref{sec: discussion} and the paper is concluded with Section \ref{sec: conclusions}.

\section{Background} \label{sec: background}

\subsection{Ising model}

This paper revolves around Oscillatory Neural Networks (ONNs), an emerging physics-based paradigm rooted in the dynamics of coupled oscillators \cite{todri2024computing}. In contrast to Artificial Neural Networks \cite{rosenblatt1958perceptron, yegnanarayana2009artificial, zou2009overview}, ONN computing boils down to finding global minima in an underlying energy function \cite{izhikevich2003simple, wang2019oim, todri2024computing}. The ONNs are inherently linked to the Ising model \cite{Ising_1925, lucas2014ising}, a mathematical model of ferromagnetism describing a spin configuration of a system \cite{Ising_1925}. In particular, the dynamics of the model can be described with the following Hamiltonian:
\begin{equation} \label{ising_hamiltonian}
    H=-\sum_{(i,j)} J_{ij} \sigma_i \sigma_j.
\end{equation}
Given a coupling weight matrix $J_{ij}$, the goal lies in identifying the natural state of the system by minimizing the outlined Hamiltonian through the set of binary spins $\sigma_i$. 

\subsection{Oscillatory Neural Networks (ONNs)}
One can bridge the gap between the Ising model and ONNs by mapping the spins $\sigma_i \in \{1,-1\}$ to the oscillators' phases $\phi_i \in \{0, \pi\}$ resulting in the corresponding ONN-Hamiltonian. Moreover, the ordinary differential equation (ODE) describing the dynamics of $N$ oscillators in an ONN can thus be determined simply by following the negative gradient of the ONN-Hamiltonian \cite{Hoppensteadt_2000, wang2019oim}:
\begin{equation} \label{kuramoto_ode}
    \Dot{\phi_i}=\frac{1}{N}\sum_{j} W_{ij} \sin{(\phi_j-\phi_i)},
\end{equation}
where $W_{ij}$ stands for the coupling between the $i$-th and $j$-th oscillator. The ODE in Eq. \ref{kuramoto_ode} is often called the Kuramoto model \cite{Kuramoto_1984}. As already mentioned, the phases ($0, \pi$) encode the Booleans ($1,-1$). However, when running an ONN, it rarely happens that the phases settle at the pre-established binary values $0$ and $\pi$, but rather at continuous ones. Hence, to force the oscillators into the binary phases, one introduces an additional term scaled by a factor $K_S$ into the ODE called \emph{Subharmonic Injection Locking} (SHIL) which injects the second harmonics into the oscillators \cite{todri2024computing, cheng2024control, wang2019oim}:
\begin{equation} \label{kuramoto_ode_with_shil}
    \Dot{\phi_i}=\frac{1}{N}\sum_{j} W_{ij} \sin{(\phi_j-\phi_i)}-K_S \sin{(2\phi_i)}.
\end{equation}

\section{Methods} \label{sec: methods}

In this Section, we introduce our approach for solving Sudoku puzzles with Oscillatory Neural Networks. Since the rules of Sudoku imply non-repeating values across all columns, rows, and boxes, Graph Coloring presents a suitable approach for solving this problem. At the end of this Section, we outline the benchmarking process.

\subsection{Proposed ONN-Sudoku solver}

\subsubsection{Graph Coloring with ONNs}
Given a graph, the aim of the Graph Coloring problem boils down to assigning each node in a graph a color such that no two adjacent nodes share the same color and a minimal number of colors is utilized. Being a part of the 21 Karp NP-complete problems \cite{karp2009reducibility}, Graph Coloring is a well-known hard combinatorial optimization problem. As shown by the author in \cite{lucas2014ising}, all NP-hard problems can be mapped to a corresponding Ising Hamiltonian with Graph Coloring not being an exception.

Furthermore, authors in \cite{mallick2022computational} demonstrated the mapping for some non-binary hard optimization problems to Oscillatory Ising Machines (\say{Oscillatory Neural Networks}). In particular, Graph Coloring is defined as a Max-$K$-Cut with the minimal $K$ such that no two adjacent nodes share the same set. In the case of a Sudoku, the number of sets or $K$ is well-known and does not need to be minimized, as it corresponds to the size of the Sudoku. Thus, the desired phases $\phi_{\text{ideal}}$ for a size $K$ Sudoku read as follows:
\begin{equation} \label{eq: desired_n_phase_roots}
    \phi_{\text{ideal}} = \biggl\{\frac{2\pi (k-1)}{K} \bigg| k \in \{1,2,...,K\}\biggr\}.
\end{equation}

Hence, for a $9\times9$ Sudoku ($K=9$), we get the following mapping of the numbers to the phases:
\begin{equation}
    \begin{split}
        1 &\rightarrow 0 \\
        2 &\rightarrow \frac{2\pi}{9} \\
        3 &\rightarrow \frac{4\pi}{9} \\
        4 &\rightarrow \frac{2\pi}{3} \\
        5 &\rightarrow \frac{8\pi}{9} \\
        6 &\rightarrow \frac{10\pi}{9} \\
        7 &\rightarrow \frac{4\pi}{3} \\
        8 &\rightarrow \frac{14\pi}{9} \\
        9 &\rightarrow \frac{16\pi}{9} 
    \end{split}
\end{equation}

\begin{figure}[h]
\centering
\includegraphics[width=\linewidth]{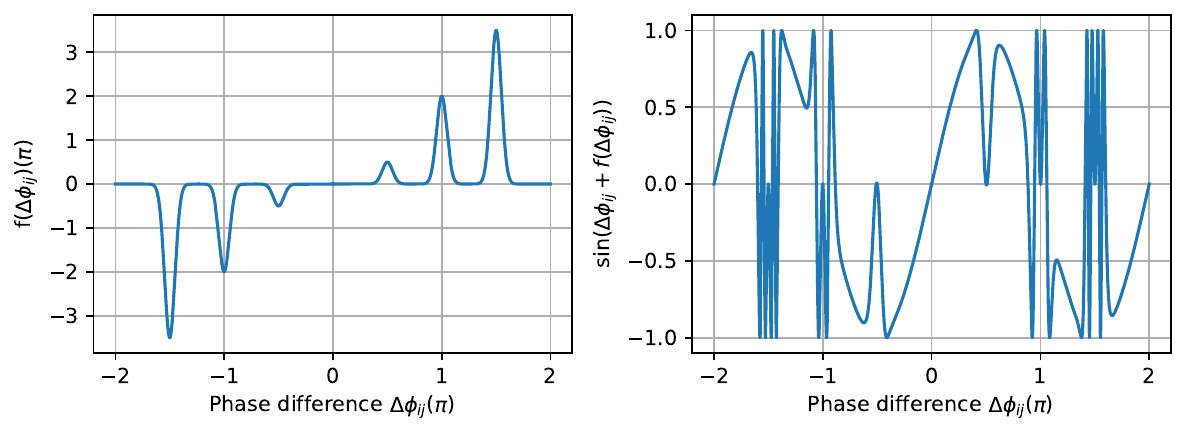}
\caption{Function $f(\Delta \phi_{ij})$ from Eq. \ref{eq:mallick_max_k_cut_f_function} (left) and term $\sin{(\Delta \phi_{ij}+f(\Delta \phi_{ij}))}$ (right) for $K=4$.}
\label{fig:f_function_investigation}
\end{figure}

Let us now turn our attention towards Max-$K$-Cut. Given a graph, the Max-Cut problem entails finding a partition between the nodes into 2 sets such that the sum of the cut edges is maximized. In the case of Max-$K$-Cut, a partition into $K$ sets with the maximum sum of the cut edges shall be found. The authors in \cite{mallick2022computational} solve Max-$K$-Cut with Oscillatory Neural Networks by extending ODE from Eq. \ref{kuramoto_ode_with_shil} to the following form:
\begin{equation} \label{eq:mallick_max_k_cut_ode}
    \begin{split}
        \Dot{\phi_i}=&-\frac{1}{N}\sum_{j} W_{ij} \sin{(\phi_i-\phi_j + f(\phi_i-\phi_j))}\\
        &-K_S \sin{(K\phi_i)},
    \end{split}
\end{equation}
with $f(\Delta \phi_{ij})$ ensuring that the ideal phases from Eq. \ref{eq: desired_n_phase_roots} produce a vanishing gradient when acted upon with a sine function. Furthermore, the function $f$ reads as follows:
\begin{equation} \label{eq:mallick_max_k_cut_f_function}
    \begin{split}
        f(\Delta \phi_{ij}) = &\sum_{k=1}^{K-1} \left((2k-1)\pi-\frac{2k\pi}{K}\right)\\
        &\biggl(\exp{-\frac{(\Delta \phi_{ij}-\frac{2k\pi}{K})^2}{2\sigma^2}} \\
        &-\exp{-\frac{(\Delta \phi_{ij}+\frac{2k\pi}{K})^2}{2\sigma^2}}\biggl),
    \end{split}
\end{equation}
with $\sigma$ being a tuneable parameter. Fig. \ref{fig:f_function_investigation} depicts the function $f(\Delta \phi_{ij})$ from Eq. \ref{eq:mallick_max_k_cut_f_function} and the term $\sin{(\Delta \phi_{ij}+f(\Delta \phi_{ij}))}$ for $K=4$ and a relatively large $\sigma$ ($=0.15$) to showcase the form of the function.

Clearly, utilizing the model from Eq. \ref{eq:mallick_max_k_cut_ode} has multiple drawbacks. First of all, the function in Eq. \ref{eq:mallick_max_k_cut_f_function} has a tuneable parameter $\sigma$, which controls the width of the peaks. Second, computing the terms in Eq. \ref{eq:mallick_max_k_cut_f_function} is computationally expensive, particularly when done so for multiple thousands of time steps and large problems. However, most importantly, the process of reaching the roots of the term $\sin{(\Delta \phi_{ij}+f(\Delta \phi_{ij}))}$ is not identical, since the gradients can vary significantly depending on the chosen route. This can lead to certain roots being preferred or in turn avoided completely. Thus, we need an easily computable model without any tuneable parameters and where each root is identical. All of these conditions are satisfied with the following model:
\begin{equation} \label{eq:correct_max_k_cut_eq}
    \Dot{\phi_i}=-\frac{1}{N}\sum_{j} W_{ij} \sin{\left(\frac{K}{2}(\phi_i-\phi_j)\right)}-K_S \sin{(K\phi_i)}.
\end{equation}
In particular, the first term ensures that the oscillators are $\frac{2\pi n}{K}$ apart, while the second term guarantees that the oscillators settle at $\frac{2\pi n}{K}$ values. We utilize Eq. \ref{eq:correct_max_k_cut_eq} instead of Eq. \ref{eq:mallick_max_k_cut_ode} in our Sudoku approach. 

Based on Eq. \ref{eq:correct_max_k_cut_eq}, we define an order parameter $\kappa$ to track how well the oscillators converged to the desired phases as outlined in Eq. \ref{eq: desired_n_phase_roots}:

\begin{equation}
    \kappa =\frac{1}{N} \sum_i 1- \left|\text{Im}\left(\exp{\left(\frac{iK\phi_i}{2}\right)}\right)\right|.
\end{equation}

The order parameter is bounded to the range $[0, 1]$, where values close to $1$ indicate that the phases $\phi_i$ converged well to the pre-established phases $\phi_{ideal}$, whereas values near $0$ hint at a poor convergence to $\phi_{ideal}$. Let us now discuss the form of the weight matrix $W_{ij}$.

\subsubsection{Weight matrix}
Let us now turn our attention to the weight matrix. First of all, we impose a zero diagonal. Second, thinking of the rules of Sudoku, there should not be any repeating values in each row, column and box. Thus, each cell should only be connected to all cells in its corresponding row, column and box. Since the Graph Coloring problem is defined as a Max-$K$-Cut with the minimal $K$ such that no two adjacent nodes share the same set, the value connecting any two cells in a common row, column or box should be negative. Without loss of generality, we opt for a value of $-1$. 

Third and most importantly, not all weights should have equal strength, as the knowns should have a stronger influence on the dynamics than the unknowns. This implies that a weight between two unknowns (let us call it $w_{u}$) shall be negative and have a smaller magnitude than the weight connecting a known ($-1$ in our case). Putting all of these conditions together, we arrive at the following form of the weight matrix $W_{ij}$:

\begin{equation} \label{eq:full_weight_matrix}
    W_{ij}=
    \begin{cases}
    -1, \text{i $\neq$ j, i $\land$ j $\in$ row, column or box, i $\lor$ j $\in$ known,}\\
    w_u, \text{i $\neq$ j, i $\land$ j $\in$ row, column or box, i $\land$ j $\notin$ known,}\\
    0, \text{else,}
    \end{cases}
\end{equation}

with the following two conditions:

\begin{itemize}
    \item $w_u<0$,
    \item $|w_u|<1$.
\end{itemize}

While the weight matrix in Eq. \ref{eq:full_weight_matrix} guarantees that the knowns will have a stronger influence on the unknowns than the unknowns, the ODE does not ensure that the unknowns will have no influence on the knowns. Hence, to produce a vanishing gradient for knowns, we modify the ODE presented in Eq. \ref{eq:correct_max_k_cut_eq} to the following form:

\begin{equation} \label{eq:correct_max_k_cut_eq_with_unknowns}
    \begin{split}
        \Dot{\phi_i}=&-f_{\text{unknown}}(\phi_i)\frac{1}{N}\sum_{j} W_{ij} \sin{\left(\frac{K}{2}(\phi_i-\phi_j)\right)} \\
        &-f_{\text{unknown}}(\phi_i)K_S \sin{(K\phi_i)},
    \end{split}
\end{equation}

with 

\begin{equation} \label{eq:unknown_function}
    f_{\text{unknown}}(\phi_i)=
    \begin{cases}
    1, \text{$\phi_i \in$ unknown,}\\
    0, \text{else.}
    \end{cases}
\end{equation}

\subsubsection{Ensuring different phases}

While the model presented in Eq. \ref{eq:correct_max_k_cut_eq_with_unknowns} accordingly describes the ONN-ODE for Max-K-Cut, it has a serious shortcoming in regards to Sudoku. Figure \ref{fig:sudoku4x4_wrong} showcases an instance of a simple 4$\times$4 Sudoku puzzle with a violation of the rules.

\begin{figure}[h]
    \centering
    \begin{tikzpicture}[scale=0.9]
      \draw[step=1cm, line width=0.3pt] (0,0) grid (4,4);
      \draw[line width=2.5pt] (0,0) rectangle (4,4);
      \draw[line width=2.5pt] (0,2) -- (4,2);
      \draw[line width=2.5pt] (2,0) -- (2,4);
    
      \foreach \x/\y/\n in {
        0.5/3.5/1, 1.5/3.5/\textbf{2}, 2.5/3.5/, 3.5/3.5/\textbf{1},
        0.5/2.5/\textbf{1}, 1.5/2.5/, 2.5/2.5/\textbf{2}, 3.5/2.5/,
        0.5/1.5/, 1.5/1.5/\textbf{3}, 2.5/1.5/, 3.5/1.5/\textbf{2},
        0.5/0.5/\textbf{2}, 1.5/0.5/, 2.5/0.5/\textbf{3}, 3.5/0.5/
      }{\node at (\x,\y) {\n};}
    \end{tikzpicture}
    \caption{Example of a 4$\times$4 Sudoku grid with a violation of the rules, where the thick numbers indicate the known values.}
    \label{fig:sudoku4x4_wrong}
\end{figure}

Clearly, the digit $1$ in the upper left hand side corner violates two rules simultaneously, as there is already a known digit $1$ in the same column as well as box. However, despite breaking the rules, the ODE presented in Eq. \ref{eq:correct_max_k_cut_eq_with_unknowns} would produce a vanishing gradient, for the digits $1$ and $1$ are indeed $\frac{2\pi n}{K}$ apart ($n=0$ is also permitted) and both values are one of the valid roots. In summary, despite the structure, the ODE presented in Eq. \ref{eq:correct_max_k_cut_eq_with_unknowns} allows the ONN to break the Sudoku rules. 

In order to circumvent this issue, we need an additional term in the ODE. This new term shall kick the oscillators out of a stable state if a violation of the rules is detected. The ODE of the modified proposed approach to solve Sudoku with Max-K-Cut reads as follows:

\begin{equation} \label{eq:correct_sudoku_eq}
    \begin{split}
        \Dot{\phi_i}=&-f_{\text{unknown}}(\phi_i)\frac{1}{N}\sum_{j} W_{ij} \sin{\left(\frac{K}{2}(\phi_i-\phi_j)\right)}\\
        &-f_{\text{unknown}}(\phi_i)K_S \sin{(K\phi_i)}\\
        &+f_{\text{unknown}}(\phi_i)K_G \sum_j W_{ij}^{'} g(\phi_i, \phi_j),
    \end{split}
\end{equation}

with $K_G$ being a scaling factor, $W_{ij}^{'}$ a weight matrix that does not differentiate between knowns and unknowns, i.e., 

\begin{equation}
    W_{ij}^{'}=
    \begin{cases}
    -1, \text{i $\neq$ j, i $\land$ j $\in$ row, column or box,}\\
    0, \text{else,}
    \end{cases}
\end{equation}

and $g(\phi_i, \phi_j)$ producing a non-vanishing value if and only if $\phi_i$ and $\phi_j$ are the $n$ multiples of $2\pi$ far apart from each (i.e. they output the same digit) or in mathematical terms:

\begin{equation}
    g(\phi_i, \phi_j)=
    \begin{cases}
    1, |\phi_i-\phi_j|\approx2\pi n,\\
    0, \text{else.}
    \end{cases}
\end{equation}

We emulate $g(\phi_i, \phi_j)$ with the following function:

\begin{equation} \label{eq:our_g_function}
    g(\phi_i, \phi_j) = \exp{\left(-\frac{(1-\cos{(\phi_i-\phi_j)})}{\sigma}\right)},
\end{equation}

where $\sigma$ tunes the width of the peaks around the $n$-th multiples of $2\pi$. 

\subsection{Benchmark set-up}

The model presented in Eq. \ref{eq:correct_sudoku_eq} has four tuneable parameters, namely:

\begin{itemize}
    \item the weight between two unknowns $w_{u}$,
    \item the scaling factor of the $K$-th harmonics $K_S$,
    \item the scaling factor of the $g(\phi_i, \phi_j)$ function $K_G$,
    \item the width of the $g(\phi_i, \phi_j)$ function's peaks $\sigma$. 
\end{itemize}

Given the amount of tuneable parameters, a thorough sweep must first be conducted to identify the appropriate combinations of parameters. Hence, we generate a training set with a size of 250 samples to narrow down the space of parameter combinations and finally utilize a test set with a size of 1000 samples to report on the best achieved performance. The Sudoku puzzles are created with the Py-Sudoku library \cite{pysudoku2024}. In particular, we benchmark our proposed ONN-Suduku solver (see Eq. \ref{eq:correct_sudoku_eq}) on Sudoku puzzles of sizes $4\times4$ and $9\times9$ for seven different unknown ratios (i.e. how many of the digits in the puzzle are not known) $[0.125, 0.25, 0.375, 0.5, 0.625, 0.75, 0.875]$ against an HNN Sudoku-solver \cite{mladenov2011solving} and an already established ONN Sudoku-solver \cite{haverkort2025solving} in terms of accuracy. Finally, we report the order parameters recorded for each unknown ratio and each size to assess how well the model converged towards the pre-established roots. All benchmarks are performed in Python.

\begin{figure*}[t]
\centering
\begin{subfigure}{.5\linewidth}
  \centering
  \includegraphics[width=\linewidth]{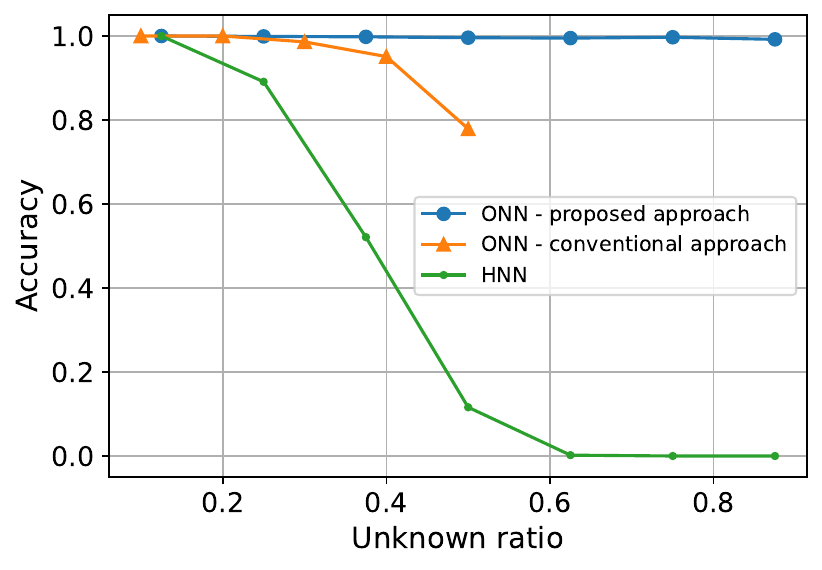}
  \caption{ }
  \label{fig:4x4_accuracy}
\end{subfigure}%
\begin{subfigure}{.5\linewidth}
  \centering
  \includegraphics[width=\linewidth]{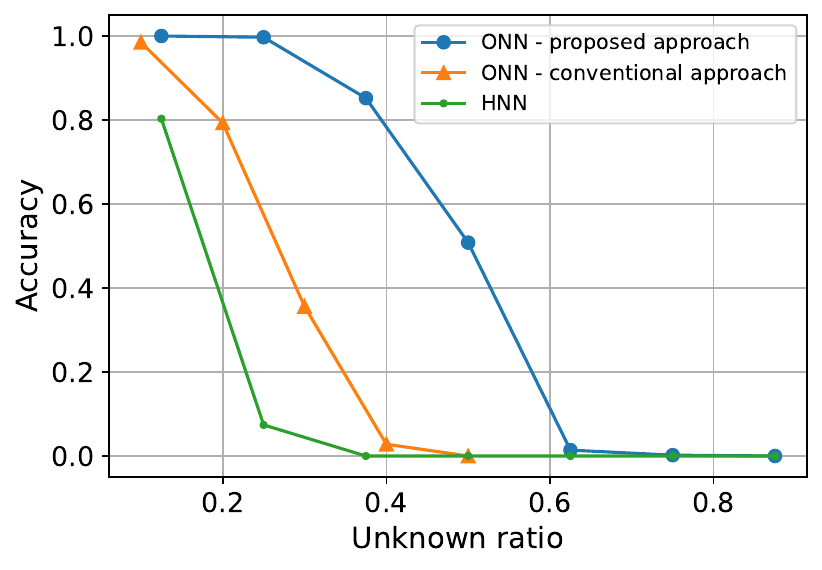}
  \caption{ }
  \label{fig:9x9_accuracy}
\end{subfigure}
\caption{Accuracy as a function of unknown ratio of the HNN-Sudoku solver \cite{mladenov2011solving}, the conventional ONN-Sudoku solver \cite{haverkort2025solving} and our proposed ONN-Sudoku solver for (a) $4\times4$ and (b) $9\times9$ Sudoku puzzles, respectively. For our ONN- and the HNN-Sudoku solver, we present the accuracies achieved on 1000 samples and for the conventional ONN approach, we plot the accuracy values reported in \cite{haverkort2025solving}.}
\label{fig:accuracy_plots}
\end{figure*}

\section{Results} \label{sec: results}

First, we start with the accuracy benchmarks for $4\times4$ and $9\times9$ Sudoku puzzles. In particular, we report on the best accuracies achieved with our ONN- and the HNN-Sudoku solver \cite{mladenov2011solving} for seven different unknown ratios $[0.125, 0.25, 0.375, 0.5, 0.625, 0.75, 0.875]$ on 1000 Sudoku puzzles. Additionally, we include the accuracies of the already developed ONN-Sudoku solver for five different unknown ratios $[0.1, 0.2 , 0.3, 0.4, 0.5]$ from the authors in \cite{haverkort2025solving}. Figures \ref{fig:4x4_accuracy} and \ref{fig:9x9_accuracy} summarize all these accuracies as a function of the unknown ratio for sizes $4\times4$ and $9\times9$, respectively.

Let us start by analyzing the results for the size $4\times4$. Regarding the HNN-Sudoku solver, as the unknown ratio increases, performance rapidly deteriorates, effectively being unable to solve any puzzles correctly for unknown ratios greater than $60\%$. The accuracy drastically improves with the conventional ONN-Sudoku solver achieving almost $80\%$ for an unknown ratio of $50\%$. In contrast, our proposed approach significantly outperforms both established methods, since it remains almost flawless throughout the benchmarks with the accuracy slightly decreasing for an unknown ratio of $90\%$. We speculate that this vastly superior performance can be traced back to the additional term presented in Eq. \ref{eq:correct_sudoku_eq} as it ensures a steady state if and only if all cells are correctly filled out.

Let us now move on to the benchmarks on $9\times9$ Sudoku puzzles. Fully analogously to previously reported benchmarks, the accuracy of the HNN solver quickly worsens as the number of unknown digits grows, effectively being incapable of solving any puzzles correctly for unknown ratios greater than $40\%$. While the conventional ONN-Sudoku solver starts off with an almost perfect accuracy, the accuracy rapidly decreases, reporting no solved puzzles for an unknown ratio of $50\%$. As far as our proposed ONN-Sudoku solver is concerned, while still producing superior accuracies in comparison to the two other methods, it is no longer able to maintain a flawless accuracy as the number of unknowns ramps up. In particular, although perfect accuracies are reached for unknown ratios up to $25\%$ and an accuracy of more than $80\%$ is obtained for an unknown ratio of $37.5\%$, the precision starts to decrease, achieving approximately $50\%$ for an unknown ratio of $50\%$ and reaching the limits of the model at unknown ratios greater than $70\%$ by not being able to solve any puzzles correctly. We assume that the order parameters will decrease as the number of unknowns increases, implying that either the additional term presented in Eq. \ref{eq:correct_sudoku_eq} has not a strong enough influence and/or the model needs more oscillatory cycles to reach a stable state.

\begin{figure*}[t]
\centering
\begin{subfigure}{\linewidth}
  \centering
  \includegraphics[width=\linewidth]{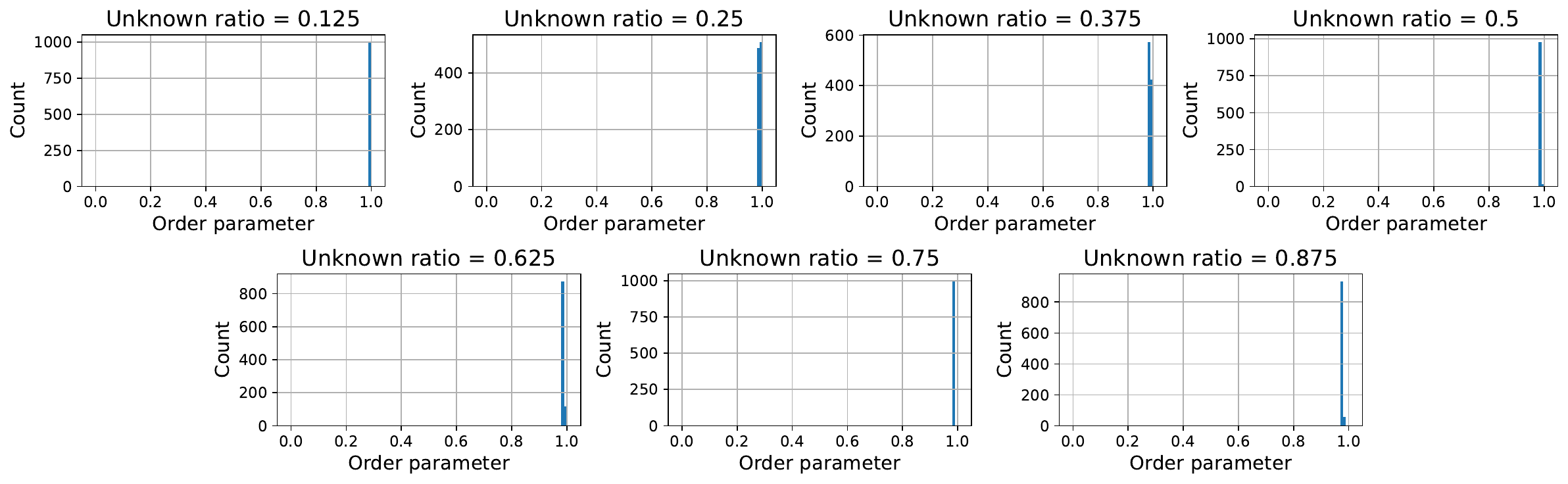}
  \caption{ }
  \label{fig:4x4_order_parameter}
\end{subfigure}%
\hfill
\begin{subfigure}{\linewidth}
  \centering
  \includegraphics[width=\linewidth]{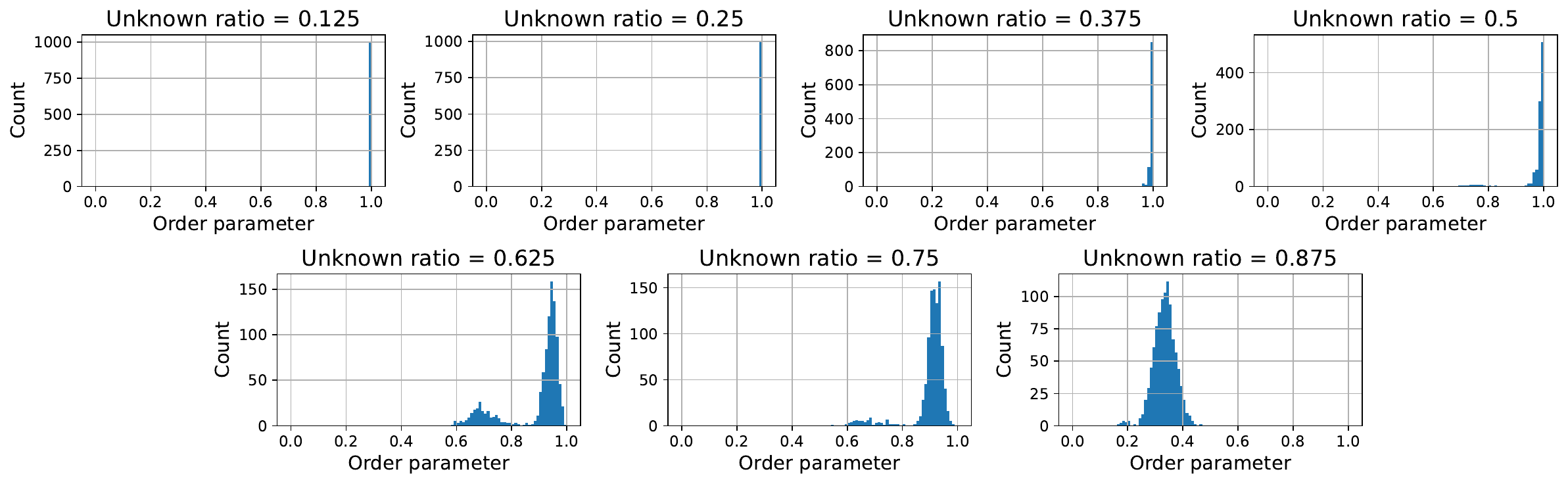}
  \caption{ }
  \label{fig:9x9_order_parameter}
\end{subfigure}
\caption{Order parameters associated with the best accuracies achieved by our proposed ONN-Sudoku solver plotted as histograms for seven different unknown ratios $[0.125, 0.25, 0.375, 0.5, 0.625, 0.75, 0.875]$ as well as  (a) $4\times4$ and (b) $9\times9$ Sudoku puzzles, respectively.}
\label{fig:order_parameter_plots}
\end{figure*}

In the second part of the benchmarks, let us closely examine the order parameters of our proposed ONN-Sudoku solver for $4\times4$ and $9\times9$ Sudoku puzzles. In particular, we report on the order parameters associated with the best accuracies reached by our solver for seven different unknown ratios $[0.125, 0.25, 0.375, 0.5, 0.625, 0.7, 0.875]$ on 1000 Sudoku puzzles. Figures \ref{fig:4x4_order_parameter} and \ref{fig:9x9_order_parameter} summarize all these order parameters as histograms for sizes $4\times4$ and $9\times9$, respectively.

Clearly, for $4\times4$ Sudokus, where almost perfect accuracies are achieved, the oscillators are able to converge consistently to their pre-established roots for all seven different unknown ratios. In contrast, for $9\times9$ Sudokus, as the number of unknown digits ramps up, the distribution of order parameters starts shifting towards lower values. In particular, while for an unknown ratio of $0.125$, almost all order parameters are clustered around $1$, for an unknown ratio of $0.875$, the vast majority of order parameters is below $0.5$. As theorized earlier, a high accuracy induces a high order parameter, while lower accuracies are associated with lower order parameters. This relationship can be attributed to the $g(\phi_i, \phi_j)$ function in Eq. \ref{eq:correct_sudoku_eq}, as it ensures a stable state if and only if no Sudoku rules are violated. A low order parameter may hint at either the additional term not being scaled appropriately and/or the ONN requiring more oscillatory cycles to reach a stable state.

\section{Discussion} \label{sec: discussion}
In this paper, we present our developed ONN-Sudoku solver inspired by the Graph Coloring problem. In Python, we successfully demonstrated that an appropriate combination of tuneable parameters can lead to almost flawless precisions for $4\times4$ Sudokus and significantly improved accuracies for $9\times9$ Sudokus. However, it needs to be acknowledged that the model we considered (see Eq. \ref{eq:correct_sudoku_eq}) has a few drawbacks. In particular, the model presented in this paper has four tuneable parameters, thus narrowing the search down through a parameter sweep to a suitable combination of parameters can be quite costly and time intensive. Due to hardware restrictions and the parameter sweep, we opted for a training set size of only 250 samples. Hence, future versions of this Sudoku solver should look into eliminating a few of the tuneable parameters. 

Furthermore, while our proposed Sudoku solver outperformed the existing HNN- and ONN-solvers, it started hitting its limits for $9\times9$ and unknown ratios larger than $50\%$. As the order parameter histograms confirmed, there is a clear relationship between accuracy and order parameter, as high accuracies induce high order parameters and low accuracies occur only with low order parameters. Based on these results, we speculate that the additional term with the $g(\phi_i, \phi_j)$ function (see Eq. \ref{eq:correct_sudoku_eq}) is not scaled properly and/or the ONN did not have enough oscillatory cycles to settle. Additionally, it may be worth looking into alternative $g(\phi_i, \phi_j)$ function formulations than the one presented in Eq. \ref{eq:our_g_function}.

While our model deploys simple, sinusoidal Kuramoto-ONNs \cite{Kuramoto_1984,wang2019oim}, there has recently been a push in the community to harness the dynamics of highly nonlinear oscillators for computing \cite{sabo2026harnessing, shougat2021information, maslennikov2025binary}. Although it is not a limitation per se, it would be worth investigating, how a highly nonlinear or even noisy oscillator would impact the accuracy as well as the order parameter compared to the noiseless Kuramoto model deployed here \cite{van_der_pol_1926, FITZHUGH1961445, 4066548, izhikevich2003simple, kovacic2011duffing, panteley2015stability}. 

\section{Conclusions} \label{sec: conclusions}
In this paper, we propose a Sudoku solver based on Oscillatory Neural Networks (ONNs) by formulating the problem as a Graph Coloring problem. Sudokus embody a well-established logical puzzle, which can be seen as a constrained combinatorial optimization problem (COP). In particular, as the Sudoku rules forbid a placement of identical digits in common rows, columns, and boxes, Graph Coloring, a problem which boils down to assigning \say{colors} to graph nodes such that no two adjacent nodes possess the same color and the number of unique colors is minimized, presents a natural COP fit.

Oscillatory Neural Networks (ONNs) present an attractive physics-based paradigm for solving hard combinatorial optimization problems by harnessing the rich dynamics of coupled oscillators. While there exists a quite computationally expensive Graph Coloring mapping to ONNs, it cannot simply be utilized for Sudokus as it allows potential rule violations. Instead, our ONN-Sudoku solver expands a modified, computationally cheaper version of Graph Coloring by an additional term which ensures that no two oscillators sharing a row, column or box yield identical digits. 

We benchmark our ONN-Sudoku solver for seven different numbers of unknown digits on $4\times4$ and $9\times9$ puzzles in terms of accuracy and order parameter against two solvers - a solver based on Hopfield Neural Networks and an already established ONN-solver. The accuracy measurements prove that our proposed solver significantly outperforms both existent solvers on $4\times4$ as well as $9\times9$ puzzles, yielding almost flawless accuracies for $4\times4$ Sudokus, while hitting a limit on $9\times9$ Sudokus for large unknown ratios. The order parameter benchmarks reveal that high accuracies are associated with large order parameter values, whereas degrading precisions occur with lower order parameters, hinting at either the additional term not being scaled properly and/or the model requiring more oscillatory cycles to stabilize.

This paper demonstrates the relevance of expanding the ODE by an additional term in ONNs for constrained combinatorial optimization to kick the oscillators out of undesired states and produce vanishing gradients for only configurations satisfying all the constraints. In particular, this term allowed ONNs to outperform already established Sudoku solvers based on ONNs and HNNs in terms of accuracy. Last but certainly not least, to improve the performance, future work should investigate the benefits of highly nonlinear oscillators and different additional term function formulations.

\bibliographystyle{IEEEtran}
\bibliography{References}

\end{document}